\documentclass[10pt,twocolumn,letterpaper]{article}

\usepackage{iccv}
\usepackage{times}
\usepackage{epsfig}
\usepackage{graphicx}
\usepackage{amsmath}
\usepackage{amssymb}

\usepackage{multirow}
\usepackage[pagebackref=true,breaklinks=true,letterpaper=true,colorlinks,bookmarks=false]{hyperref}
\usepackage[title]{appendix}

\iccvfinalcopy 

\def\iccvPaperID{****} 

\ificcvfinal\fi

\begin{document}

\title{Suppress-and-Refine Framework for End-to-End 3D Object Detection}

\author{
Zili Liu\textsuperscript{\rm 1}\thanks{Equal contribution} \quad
Guodong Xu\textsuperscript{\rm 1*} \quad
Honghui Yang\textsuperscript{\rm 1}\quad
Minghao Chen\textsuperscript{\rm 1} \\
Kuoliang Wu\textsuperscript{\rm 1} \quad
Zheng Yang\textsuperscript{\rm 2} \quad
Haifeng Liu\textsuperscript{\rm 1} \quad
Deng Cai\textsuperscript{\rm 1,2}\thanks{Corresponding author} \\
\textsuperscript{\rm 1}State Key Lab of CAD\&CG, Zhejiang University, Hangzhou \\
\textsuperscript{\rm 2}Fabu Inc., Hangzhou \\
{\tt\small \{zililiuzju, minghaochen01, wukuoliangzju\}@gmail.com \quad yangzheng@fabu.ai} \\
{\tt\small \{memoiry, yanghonghui, haifengliu, dcai\}@zju.edu.cn}
}

\maketitle
\ificcvfinal\thispagestyle{empty}\fi

\begin{abstract}
   3D object detector based on Hough voting achieves great success and derives many follow-up works. Despite constantly refreshing the detection accuracy, these works suffer from handcrafted components used to eliminate redundant boxes, and thus are non-end-to-end and time-consuming. In this work, we propose a \textbf{suppress-and-refine} framework to remove these handcrafted components. To fully utilize full-resolution information and achieve real-time speed, it directly consumes feature points and redundant 3D proposals. Specifically, it first \textbf{suppresses} noisy 3D feature points and then feeds them to 3D proposals for the following RoI-aware \textbf{refinement}. With the gating mechanism to build fine proposal features and the self-attention mechanism to model relationships, our method can produce high-quality predictions with a small computation budget in an end-to-end manner. To this end, we present the first fully end-to-end 3D detector, SRDet, on the basis of VoteNet. It achieves state-of-the-art performance on the challenging ScanNetV2 and SUN RGB-D datasets with the fastest speed ever. Our code will be available at \url{https://github.com/ZJULearning/SRDet}.
\end{abstract}

\section{Introduction}

3D object detection aims to localize and recognize 3D objects in a 3D scene. A large number of practical applications rely on 3D object detection, such as augmented reality \cite{park2008multiple}, robot navigation \cite{gomez2016pl,geiger2012we}, robot grasping \cite{ramon2017detection}, etc. At present, 3D object detection has become a popular research topic and has received widespread attention with the popularity of depth sensors.
 
In the 3D object detection area, early work proposes VoteNet \cite{qi2019deep} based on Hough voting, which achieves promising performance. Subsequently, a lot of progress has been made in follow-up works \cite{zhang2020h3dnet,chen2020hierarchical,xie2020mlcvnet,DBLP:journals/corr/abs-2104-06114,DBLP:journals/corr/abs-2104-00678}. However, these 3D detectors still suffer from complicated and time-consuming handcrafted components, e.g., non-maximum suppression (NMS). As shown in Table \ref{speed}, the NMS costs $\geq 20$ms in all these detectors, which severely reduces efficiency and affects deployment. 

To address this NMS-related issue, adopting the self-attention mechanism to model relationships between proposals and the bipartite matching strategy to supervise training have recently become popular in the 2D domain \cite{carion2020end,DBLP:journals/corr/abs-2010-04159,DBLP:journals/corr/abs-2012-05780,DBLP:journals/corr/abs-2011-12450}. It allows detectors to predict all objects at once without any handcrafted components. However, these methods cannot be directly applied to the 3D domain for two reasons: geometric point noises caused by the disordered and varying-density nature of point clouds and the substantial computation overheads required by the 3D space.

\begin{table}[tbp]
\centering
\resizebox{\columnwidth}!{\begin{tabular}{l|c|cc|c}
\hline
\multirow{2}{*}{Method} & \multirow{2}{*}{mAP@0.5} & \multicolumn{3}{c}{Time} \\
\cline{3-5} 
& & Model & NMS & Total  \\
\hline
VoteNet* \cite{qi2019deep} & 40.1 & 50 & 43 & 93 \\
H3DNet \cite{zhang2020h3dnet} & 48.1 & 204 & 23 & 227 \\
MLCVNet \cite{xie2020mlcvnet} & 41.4 & 53 & 96 & 149 \\
BRNet \cite{DBLP:journals/corr/abs-2104-06114} & 50.9 & 56 & 59 & 115 \\
Group-Free \cite{DBLP:journals/corr/abs-2104-00678} & 48.9 & 99 &42 & 141 \\
\hline
Ours (Refine 1 time) & 50.1 & 63 & 0 & 63 \\
Ours (Refine 3 times) & 53.5 & 74 & 0 & 74 \\
\hline
\end{tabular}}
\caption{Comparison of accuracy and average inference time (ms) on ScanNetV2 validation set. The inference time is measured on a single NVIDIA Tesla V100. * denotes that the model is re-implemented by MMDetection3D.}
\label{speed}
\end{table}

\begin{figure*}[!t]
\centering
\includegraphics[width=2\columnwidth]{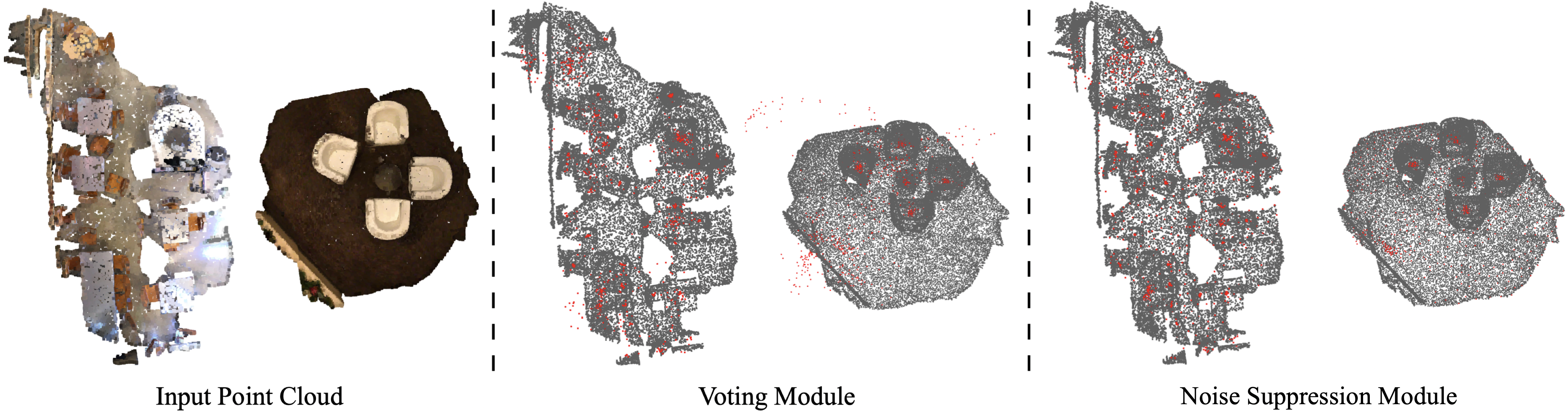}
\caption{Illustration of vote points (red dots). In the voting module, many vote points appear to be randomly distributed or scattered outside the scene. After adopting the noise suppression module, those uncontrollable vote points are well suppressed.}
\label{harm}
\end{figure*}

In this work, we design a 3D suppress-and-refine framework to remove those handcrafted components on the basis of VoteNet, which solves the noise and overhead problems at once. The suppress-and-refine framework regards redundant boxes predicted by VoteNet as initial proposals and refines these proposals using noise-suppressed feature points in a RoI-aware manner. It not only can produce precise detection results, but also can identify redundant proposals with the help of the self-attention mechanism. 

Specifically, the suppress-and-refine framework consists of the noise suppression module and the proposal refinement module. The noise suppression module restricts inaccurate geometric feature points for the subsequent RoI-aware refinement that is sensitive to point noises. As shown in Figure \ref{harm}, many feature points from VoteNet \cite{qi2019deep} (i.e., vote points) appear to be randomly distributed or even scattered outside the scene. Our experiments illustrate that those scattered points are very likely to come from uncontrollable offset predictions of background seed points. To address this issue, we propose this module to restrict these noisy offset predictions of background seed.

In the proposal refinement module, we propose the multi-resolution RoI pooling and effective gating mechanism to construct rich proposal features while concerning 3D visual characteristics and computational overheads. With the self-attention mechanism and bipartite matching strategy, this module can explicitly identify redundant proposals and thus turn the VoteNet into a fully end-to-end 3D detector. Besides, the module can be used iteratively to refine the detection results to achieve better performances.





With the suppress-and-refine framework, we propose a novel end-to-end 3D object detector named SRDet. In our experiments, SRDet achieves a state-of-the-art performance on popular challenging 3D object detection datasets: ScanNetV2 \cite{dai2017scannet} and SUN RGB-D \cite{song2015sun}. Besides, its inference speed is faster than existing 3D object detectors. Our contribution can be summarized as follows:

\begin{itemize}
\item We introduce a novel suppress-and-refine framework to efficiently refine 3D proposals using noise-suppressed feature points and identify redundant proposals in an end-to-end manner. 

\item We show that VoteNet produces scattered vote points due to uncontrollable offset predictions of background seed points, and suppressing these offset predictions is beneficial for reducing noisy vote points. 

\item We extract RoI features and use a gating mechanism to obtain fine proposal features with the 3D visual characteristic and computational overhead considered.

\item The proposed framework well fits widely-used VoteNet and results in the first fully end-to-end 3D object detector with the best runtime-quality trade-offs. 
\end{itemize}


\begin{figure*}[!t]
\centering
\includegraphics[width=2\columnwidth]{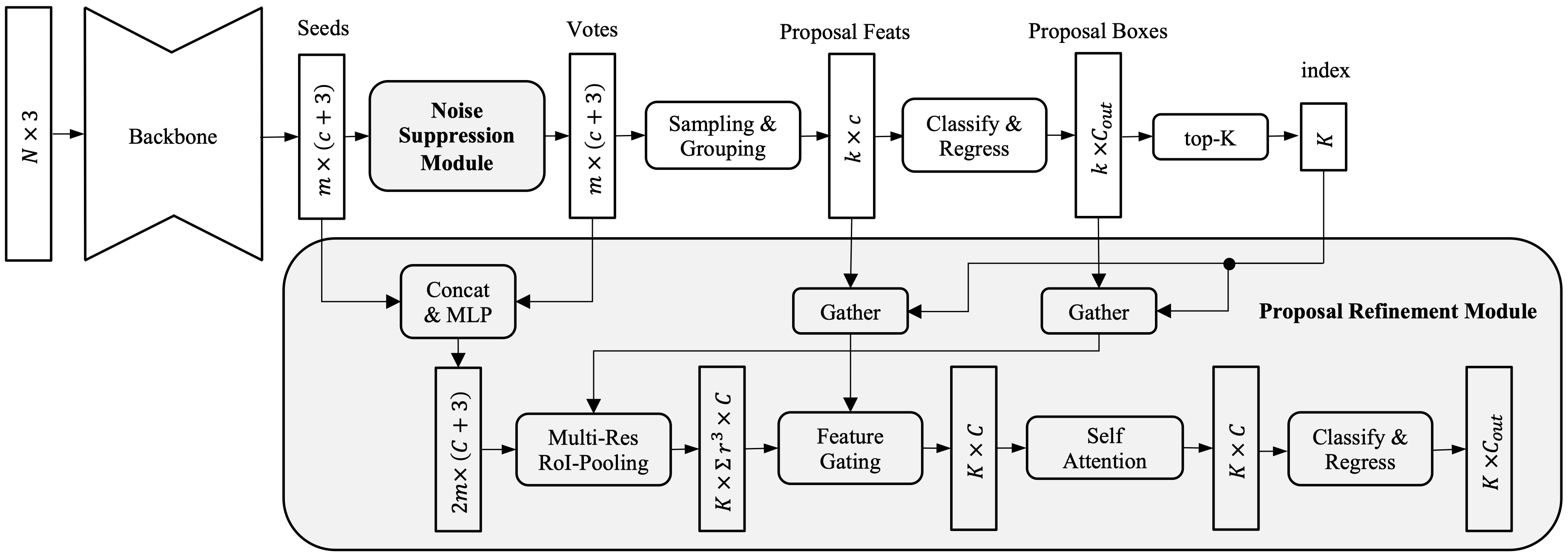}
\caption{Architecture of SRDet. SRDet consists of three components --- VoteNet, the noise suppression module, and the proposal refinement module. The noise suppression module replaces the voting module in VoteNet to produce finer votes, and the proposal refinement module aims to  produce non-redundant predictions without any handcrafted components.}
\label{stru}
\end{figure*}

\section{Related Works}

\paragraph{3D Object Detection Based on Point Cloud.} The performance of 3D object detectors continues to make breakthroughs with the development of depth sensors and deep learning. Due to the difficulty of multi-sensor fusion, mainstream 3D object detectors directly detect 3D objects on 3D point clouds and have made great progress. According to the point cloud feature extraction networks, these detectors can be divided into two streams: grid-based \cite{Yang_2018_CVPR,DBLP:journals/sensors/YanML18,DBLP:conf/cvpr/LangVCZYB19} and point-based \cite{DBLP:conf/nips/QiYSG17,shi2019pointrcnn,DBLP:conf/cvpr/YangS0J20}. Grid-based methods will first partition point clouds and use 2D CNN or 3D CNN to extract grid-level features. Then, they will take off-the-shelf 2D detection heads for proposal generation. 

Point-based methods take raw point cloud feature extraction network such as PointNet++ \cite{DBLP:conf/nips/QiYSG17} to generate point-wise features for the subsequent detection. PointRCNN \cite{shi2019pointrcnn} proposes to use a two-stage network similar to Faster-RCNN \cite{DBLP:journals/corr/RenHG015} to localize 3D objects. VoteNet \cite{qi2019deep} introduces deep hough voting to accurately gather point clouds inside objects for object-level detection. Recently, there is a popular stream attempting to combine features learned from different point cloud representations to obtain more robust features. MVF \cite{DBLP:conf/corl/ZhouSZAGOGNV19} uses the original point cloud as a bridge to integrate grid-level features in the orthogonal coordinate system and spherical coordinate system. PV-RCNN \cite{Shi_2020_CVPR} combines grid-level and point-level features to obtain stronger point cloud feature representation. Without exception, these methods rely on the one-to-many match strategy between predictions and ground-truth for the loss calculation during model training, which makes it necessary to add handcrafted components (e.g., NMS) at the end of the model to remove redundant predictions.

\paragraph{2D Object Detection.} Object detectors have been developed for several years with significant improvements. In spite of the outstanding achievements in performance \cite{DBLP:journals/corr/RenHG015,DBLP:journals/corr/abs-1712-00726,lin2017focal,tian2019fcos}, the reliance on handcrafted components makes the detector non-end-to-end and time consuming. Although some works \cite{DBLP:journals/corr/abs-1904-07850,liu2020training,LIU202159} successfully remove handcrafted components, they have limitations on the model structure. Recently, adopting self-attention mechanism to model relationships between proposals and the bipartite matching strategy to supervise training becomes popular \cite{carion2020end,DBLP:journals/corr/abs-2010-04159,DBLP:journals/corr/abs-2012-05780,DBLP:journals/corr/abs-2011-12450}. They only assign one positive sample for each ground truth, which explicitly makes detectors produce unique prediction for each object in a learnable manner.

\section{SRDet}

This section presents the proposed architecture of SRDet. As shown in Figure \ref{stru}, SRDet consists of three components: VoteNet \cite{qi2019deep} for initial proposals and point feature generation, noise suppression module to suppress noisy feature points from VoteNet, and the proposal refinement module to refine and identify redundant 3D proposals. We first briefly introduce the details of VoteNet for a background warm-up and then introduce how we extend it to SRDet with several proposed components.

\subsection{VoteNet}

Before diving into the details of the proposed modules, we will first introduce a typical 3D object detector based on Hough voting, VoteNet \cite{qi2019deep}, which is the basis of the SRDet. VoteNet mainly contains three blocks: feature extraction, voting, and proposal generation.

In feature extraction, it uses the modified PointNet++ \cite{DBLP:conf/nips/QiYSG17} as the backbone network to extract features of the point cloud directly. The backbone network consists of multiple set-abstraction layers and up-sampling layers. The output of the backbone network is denoted as $m$ seeds of dimension $c+3$, where $c$ denotes the feature dimension, and $3$ is the three-dimension coordinate of a seed point.

The point cloud is usually distributed on the surface of objects, which may be far from the object centroid. To make training easier, VoteNet proposes the voting module for centroid offset prediction. Specifically, given $m$ seeds, it learns the euclidean offset $\Delta \hat X\in R^{m\times 3}$ from seed points $X\in R^{m\times 3}$ to their corresponding object centroids. It defines vote points as $Y=X+\Delta \hat X$, which are expected to be clustered near object centroids.

After obtaining $m$ votes, VoteNet samples a subset of $K$ votes using farthest point sampling \cite{623193} and aggregates information from $m$ votes to the sampled $K$ votes. These $K$ votes are regarded as $K$ proposals for classification and regression. Finally, the NMS is applied to remove redundant proposals. In SRDet, we remove the final time-consuming NMS in VoteNet.


\subsection{Noise Suppression Module} 

\begin{figure}[!t]
\centering
\includegraphics[width=1\columnwidth]{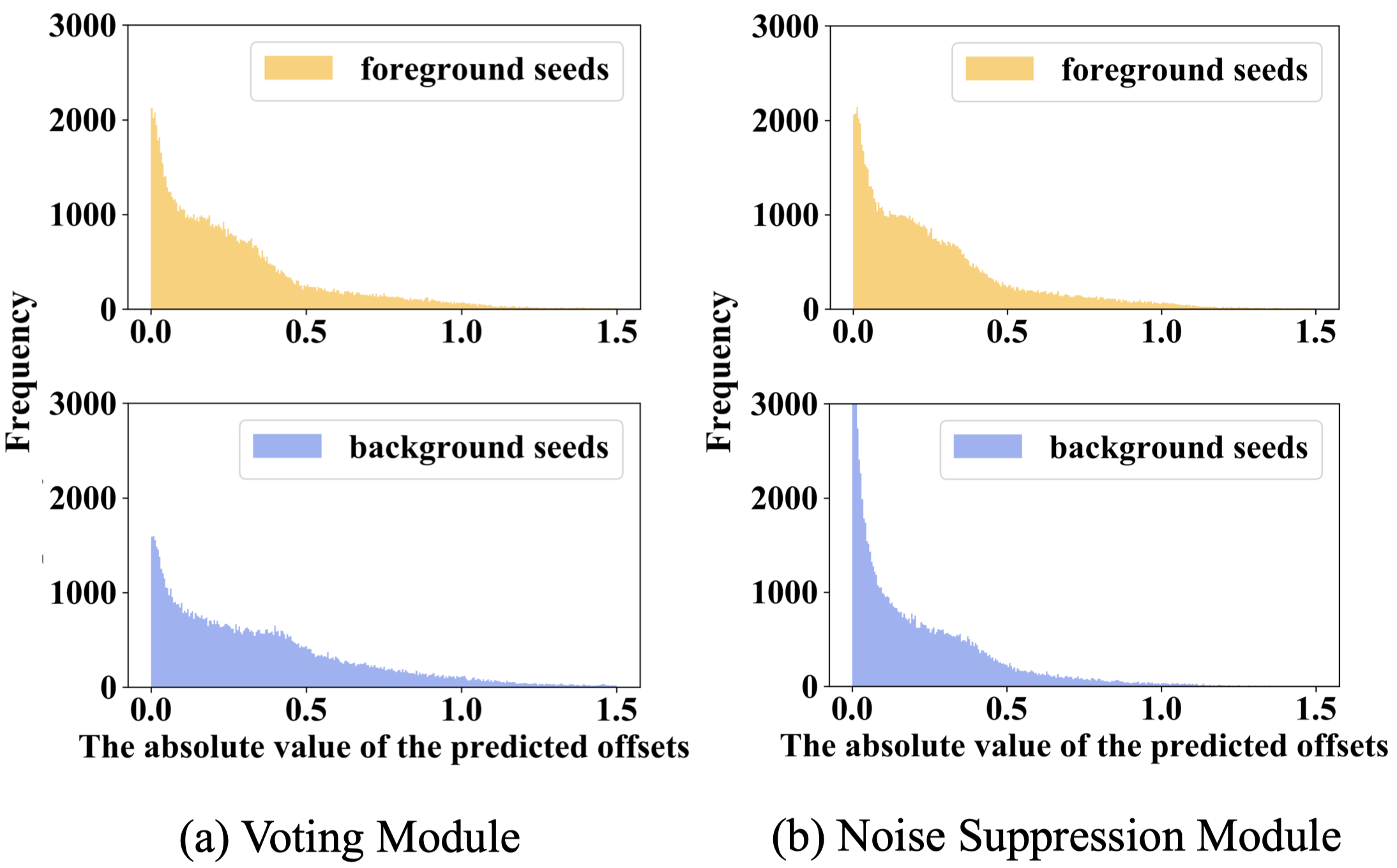}
\caption{Histogram of absolute values of offsets predicted by foreground and background seeds. We randomly select 100000 offset predictions in both (a) and (b), respectively. After adopting the noise suppression module, absolute values of offsets predicted by background seeds are reduced.}
\label{dist}
\end{figure}

In our suppress-and-refine framework, we aim at refining and identifying the redundant $K$ proposals with their geometric information and corresponding vote point features from upstream VoteNet in an end-to-end manner. We notice that many vote points are scattered outside the scene randomly instead of clustered at object centers, as shown in Figure \ref{harm}. We find that these scattered vote points mainly come from background seeds, i.e., those seed points that do not fall on any object surfaces. Since those seed points are not supervised during training, their offset predictions are uncontrollable, which is harmful to the following refinement as it may be considered as a part of proposal features. As shown in Figure \ref{dist}(a), absolute values of offsets predicted by background seeds are relatively larger than foreground seeds. It further illustrates that most of the scattered vote points come from background seeds. 

To alleviate this problem for the subsequent proposal refining, we propose a noise suppression module (NSM). It predicts a binary objectness score $\hat A=[\hat A^{-};\hat A^{+}]\in R^{m\times 2}$ to supervise all seed points. Then, we use this score to suppress offset predictions. Since background seeds are expected to have a low objectness score, their offset predictions are well suppressed. Therefore, our harmonized vote points are defined as $Y=X+ \Delta\hat{X}_p \cdot softmax(\hat A)^{+}$.

In the training phase, we define the binary objectness label $A\in R^{m\times 1}$ for $m$ seeds. Those foreground seeds located inside ground truth boxes are labeled as $1$, while background seeds are labeled as $0$. The objectness scores $\hat A$ are optimized using the cross-entropy loss $L_{vote-obj}$. Following \cite{qi2019deep}, we define regression target to supervise offset predictions and produce regression loss $L_{vote-reg}$. The difference is that our offset predictions are suppressed by the predicted objectness score. Other details about the offset predictions follow \cite{qi2019deep}.

The total loss in the noise suppression module is defined as follows, weighted by $\lambda_1$ and $\lambda_2$:
\begin{equation}
L_{nsm}=\lambda_{1}\cdot L_{vote-obj}+\lambda_{2}\cdot L_{vote-reg}
\end{equation}

\subsection{Proposal Refinement Module}

VoteNet produces $K$ prediction boxes and then applies the NMS to remove redundant parts. As shown in Table \ref{speed}, the NMS takes nearly half of the total inference time, which is inefficient. To address this issue, we use the proposal refinement module to process redundant boxes and produce a unique prediction for each object at the end. The module consists of four parts: multi-resolution RoI pooling, feature gating, self-attention, and set prediction.

\paragraph{Multi-Resolution RoI Pooling.} We reuse feature points in VoteNet to refine $K$ boxes. Specifically, we concatenate $m$ seeds and $m$ votes to form $2m$ feature points, and we extract $K$ RoI features based on the $2m$ feature points and $K$ proposal boxes using RoI pooling.

As for the pooling resolution $r\times r\times r$, its grid number is $r^3$. If $r$ is large, computational overheads will increase significantly, and there will also be many empty grids without information. Therefore, we need to keep the resolution $r$ relatively low. To capture richer features, we pool RoI features at multiple low resolutions. Specifically, we adopt three odd resolutions, i.e., 1, 3, and 5. Note that $2m$ feature points are not evenly dispersed inside the object. Instead, they are mainly located near the object boundary (i.e., seeds) and object center (i.e., votes). Intuitively, odd resolutions allow vote points located near object centers to directly fall into the center grid instead of being divided into multiple grids. After pooling, each RoI feature has the dimension of $\sum_{r\in \{1,3,5\}} r^3\times C$.

In comparison, Lidar-RCNN \cite{DBLP:journals/corr/abs-2103-15297} captures raw points located in proposal boxes and uses another backbone network for feature extraction. Their method requires separate training for the second-stage module, while our goal is to establish a fully end-to-end detector. Therefore, we reuse features instead of capturing raw point clouds and extracting features from the beginning.

\paragraph{Feature Gating.} After obtaining $K$ RoI features, we can adopt the self-attention mechanism to model relationships between proposals for identifying redundancy. However, the dimension of RoI features is too large for the widely used transformer layer. Therefore, the feature dimension needs to be reduced at first.

\begin{figure}[!t]
\centering
\includegraphics[width=0.95\columnwidth]{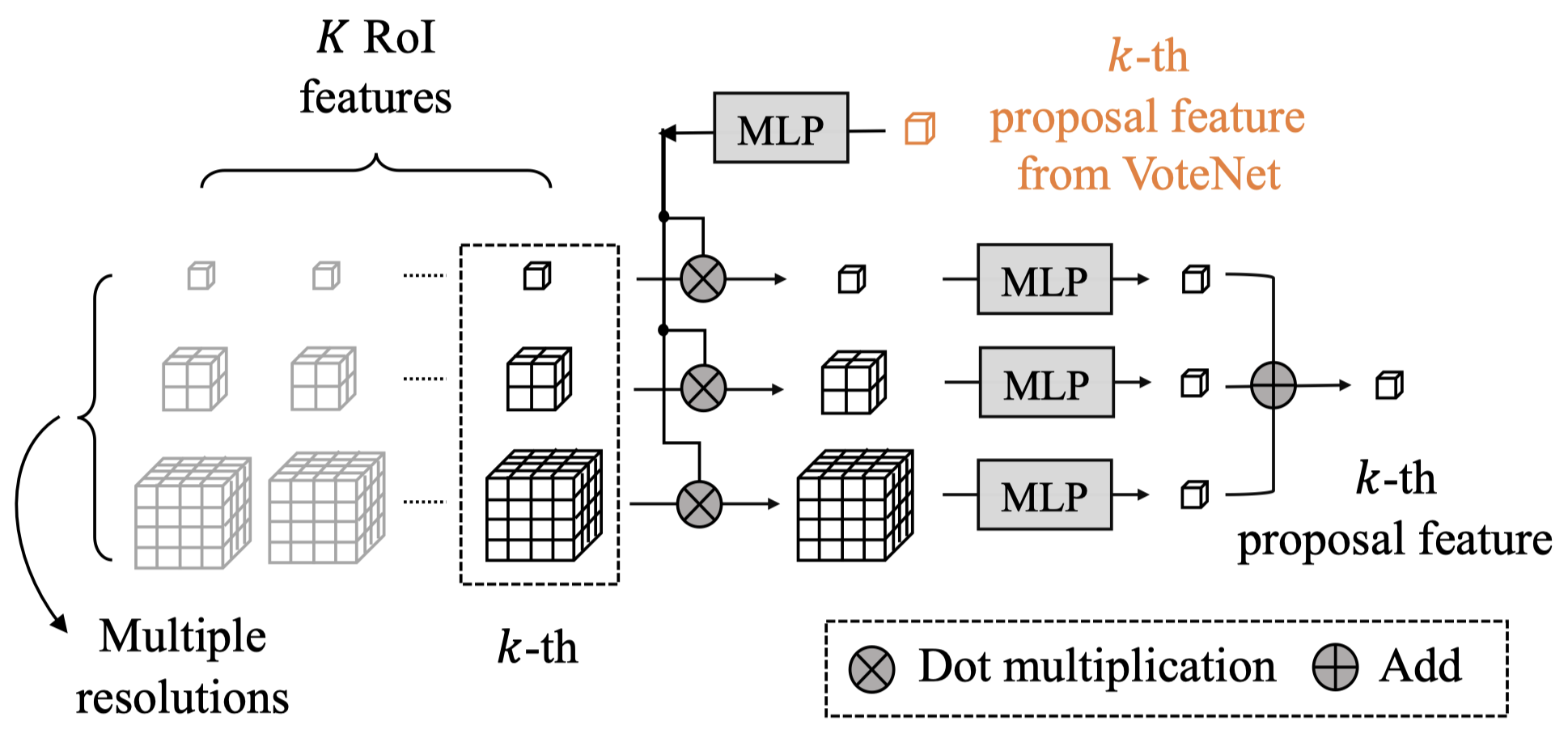}
\caption{Illustration of the feature gating. The module takes $K$ RoI features of multiple resolutions and $K$ proposal features from VoteNet as input, and it outputs $K$ proposal features. We use features from VoteNet as the gate to filter effective grids in RoI features.}
\label{inter}
\end{figure}

Since each RoI feature is composed of multiple pooled features at different resolutions, our module processes these features separately and sums them up at the end, as shown in Figure \ref{inter}. Inspired by the 2D detector Sparse R-CNN \cite{DBLP:journals/corr/abs-2011-12450}, we reuse proposal features from VoteNet as the gate to filter effective grids in RoI features. Specifically, proposal features from VoteNet are processed by a multi-layer perceptron (MLP), and then they filter RoI features of different resolutions separately. In the feature gating, we observe that dot multiplication performs better than matrix multiplication proposed in Sparse R-CNN in the 3D domain.

\paragraph{Self-Attention.} We use the widely used transformer layer \cite{carion2020end,DBLP:journals/corr/abs-2011-12450} to model relationships between $K$ proposal features. The transformer layer outputs $K$ proposal features of dimension $C$, which are used to infer $K$ vectors with semantic classification scores and box parameters (i.e., center, size, and heading angle).

\paragraph{Set Prediction.} Given $K$ predictions and $n$ ground truth, we find optimal matched $n$ predictions through the Hungarian algorithm and assign $n$ ground truth to $n$ matched predictions \cite{carion2020end}. Unmatched $K-n$ predictions are regarded as negative training samples. Our binary matching cost is defined as:
\begin{equation}
C_{match}=w_{cls} L_{cls}+w_{L1} L_{L1} + w_{iou} L_{iou}
\end{equation}

\noindent $L_{cls}$ is the focal loss \cite{lin2017focal} between predicted categories and target categories. $L_{L1}$ is the L1 loss between normalized regression results and the target results. $L_{iou}$ is the IoU loss \cite{zhou2019iou} between decoded 3D boxes and the ground truth boxes. $w_{cls}$, $w_{L1}$, $w_{iou}$ represent the weight of the focal loss, L1 loss, and IoU loss. After obtaining the optimal match, we define the same loss function $L_{prm}$ as the matching cost with an additional corner loss \cite{qi2018frustum} $L_{cor}$ weighted by $w_{cor}$ for matched pairs.

\paragraph{Multiple Refinement.} To better localize objects and make the model converge faster, we refine the prediction boxes multiple times \cite{DBLP:journals/corr/abs-1712-00726,DBLP:journals/corr/abs-2011-12450} using the same proposal refinement module. During refining iteration, we replace proposal boxes and proposal features from VoteNet with the last module output, and the module produces $K$ finer classification and regression results. We calculate the optimal matching pair and the loss $L_{prm}$ for these prediction results in each iteration, and sum all the losses as the total module loss.

The total loss $L_{total}$ is defined as: 
\begin{equation}
L_{total}=L_{votenet}+L_{nsm}+\sum_{i=1} L_{prm}^{(i)}
\end{equation}
\noindent where $L_{votenet}$ refers to the loss produced by the detection head of VoteNet, and $i$ denotes the $i$-th refinement.

\section{Experiments}
\subsection{Experimental Settings}
\paragraph{Implementation Details.} The implementation of the first-stage network follows VoteNet \cite{qi2019deep}. We remove the NMS in VoteNet and set the proposal number $k$ to $160$. Besides, we select the top-K proposals from $k$ proposals based on their scores. $K$ is $128$ in our implementation. More details about $k$ and $K$ are provided in the appendix.

The noise suppression module is implemented by a MLP layer with output channels $256$, $256$, $261$. The last layer outputs a 2-dimension objectness score, $3$-dimension centroid offset, and $256$-dimension feature offset. The loss weight $\lambda_{1}$ and $\lambda_{2}$ are set to $1.0$ and $10.0$.

In the proposal refinement module, the RoI pooling operation is implemented by the grid-level max pooling at three resolutions, i.e., resolution $r\in [1,3,5]$. $C$ is $128$, thus the full dimension of each RoI feature is $\sum_i r_i^3\times C=19584$.

The self-attention mechanism is implemented by a transformer layer. As features of VoteNet already contain position information, we do not adopt additional position embedding. The embedding dimension is $1024$, and the head number is $8$. We use the dropout layer and set the ratio to $0.1$. 

We refine prediction boxes 3 times using the proposal refinement module, and the loss weight of the module is set as $w_{cls}=1.5$, $w_{L1}=0.45$, $w_{iou}=2$, $w_{cor}=0.25$. Details about weight sensitive analysis are provided in the appendix.

\paragraph{Training Details.} Our SRDet is optimized by AdamW optimizer \cite{loshchilov2018fixing}. We train the model for $180$ epochs and the learning rate is reduced by a factor of $10$ at epoch $120$ and $160$. Besides, the weight decay is set to $0.01$. Our experiments are based on open source detection toolbox MMDetection3D \cite{DBLP:journals/corr/abs-1906-07155}.

\subsection{Comparison with State-of-the-Arts}

\paragraph{Dataset.} ScanNetV2 \cite{dai2017scannet} is a richly annotated dataset of indoor scenes. It contains about 1200 training samples and 18 object categories. ScanNetV2 does not provide orientation information of bounding boxes. Thus we predict axis-aligned boxes. In this dataset, the initial learning rate is $0.0018$, and the batch size is $16$.

SUN RGB-D \cite{song2015sun} is a single-view RGB-D dataset that contains about 5000 images annotated with oriented 3D bounding boxes. We convert RGB-D images to point clouds and follow a standard evaluation protocol, and we report results on the 10 most common categories as in VoteNet. In this dataset, the initial learning rate is $0.0016$, and the batch size is $16$.

\begin{figure}[!t]
\centering
\includegraphics[width=\columnwidth]{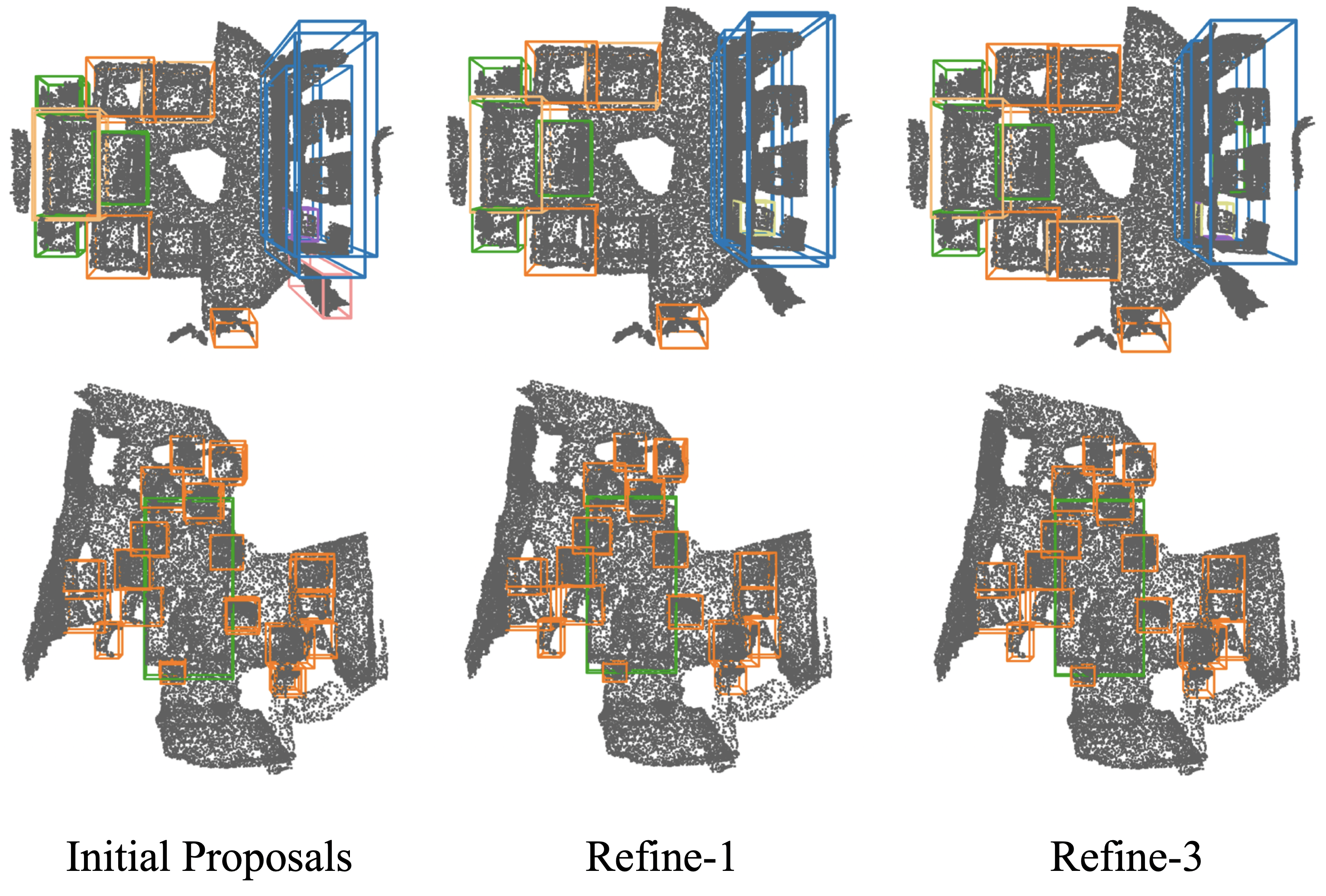}
\caption{Illustration of prediction boxes during refinement. As the refinement progresses, the model can produce finer 3D boxes while suppressing redundancy.}
\label{refine_merge}
\end{figure}

\paragraph{Results.} We summarize the results in Table \ref{results1}. The SRDet achieves 66.2 mAP@0.25 and 53.5 mAP@0.5 on ScanNetV2, 60.0 mAP@0.25 and 44.7 mAP@0.5 on SUN RGB-D. The detection accuracies are the state-of-the-art, especially for the mAP@50. It suggests that our model can locate objects accurately. Besides, as shown in Table \ref{speed}, the inference speed is faster than existing 3D object detectors.

Since we use the IoU loss and corner loss in the proposal refinement module, we adopt these losses in VoteNet for a fair comparison. Table \ref{results1} shows that although these losses bring a 4+ improvement of mAP@0.5, there is still a 5+ mAP@0.5 gap compared with the SRDet.

The recently proposed Group-Free detector \cite{DBLP:journals/corr/abs-2104-00678} achieves great performance on both ScanNetV2 and SUN RGB-D. In contrast, the SRDet has a great advantage in inference time due to the removal of the NMS. Specifically, SRDet takes 74ms to process one frame on average, while Group-Free detector takes 141ms. Besides, SRDet has fewer training epochs (i.e., 140 epochs) than Group-Free detector (i.e., 400 epochs).

\begin{table}[]
\centering
\resizebox{1\columnwidth}!{\begin{tabular}{l|cc|cc}
\hline
\multicolumn{1}{c|}{\multirow{2}{*}{Method}} & \multicolumn{2}{c|}{ScanNetV2} & \multicolumn{2}{c}{SUN RGB-D}   \\ 
\cline{2-5} 
\multicolumn{1}{c|}{} & \multicolumn{1}{l}{mAP@0.25} & \multicolumn{1}{l|}{mAP@0.5} & \multicolumn{1}{l}{mAP@0.25} & \multicolumn{1}{l}{mAP@0.5}  \\ \hline
VoteNet* \cite{qi2019deep} & 63.2 & 40.1 & 59.1 & 35.8 \\
VoteNet+IoU* & 63.5 & 44.3 & 59.7 & 39.4 \\
VoteNet+IoU+Cor* & 63.7 & 45.5 & 60.2 & 39.6 \\
HGNet \cite{chen2020hierarchical} & 61.3 & 34.4 & 61.6 & - \\
H3DNet \cite{zhang2020h3dnet} & 67.2 & 48.1 & 60.1 & 39.0 \\
MLCVNet \cite{xie2020mlcvnet} & 64.5 & 41.4 & 59.8 & -   \\ 
BRNet \cite{DBLP:journals/corr/abs-2104-06114} & 66.1& 50.9 & 61.1 & 43.7 \\
Group-Free \cite{DBLP:journals/corr/abs-2104-00678} & 67.3 & 48.9 & 63.0 & 45.2 \\
\hline
Ours & 66.2 & 53.5 & 60.0 & 44.7  \\
\hline
\end{tabular}}
\caption{Performance comparison on ScanNetV2 and SUN RGB-D. * denotes that the model is re-implemented by MMDetection3D. IoU and Cor denote the IoU loss and corner loss.}
\label{results1}
\end{table}

\subsection{Ablation Study}

In this section, we conduct our ablation study on ScanNetV2 \cite{dai2017scannet}. The results are reported on the validation dataset using a single NVIDIA Tesla V100.

\paragraph{Noise Suppression Module.} In the noise suppression module, we propose to predict objectness scores to suppress those uncontrollable offset predictions of background seeds. As shown in Figure \ref{dist}, adopting this method significantly reduces the absolute values of offsets predicted by background seeds. Besides, as shown in Figure \ref{harm}, we can see that the background points tend to stay in place instead of being randomly shifted. This avoids the noisy background vote points. Since predicting objectness scores also helps introduce more supervise information to improve the performance, we separately study the effects of supervising objectness scores and suppressing offset predictions. As shown in Table \ref{objectness}, both supervising objectness scores and suppressing offset predictions promote the performance. Specifically, they bring 1.0 and 0.7 mAP@0.25 improvements and 1.8 and 0.9 mAP@0.5 improvements, respectively. Besides, when we directly replace the voting module in VoteNet \cite{qi2019deep} with our noise suppression module, the performance improves 1.9 mAP@0.5.

\begin{table}[tbp]
\centering
\resizebox{0.85\columnwidth}!{\begin{tabular}{ccc|cc}
\hline
\multicolumn{2}{c|}{w/NSM} & \multirow{2}{*}{w/PRM} & \multirow{2}{*}{mAP@0.25} & \multirow{2}{*}{mAP@0.5} \\
\cline{1-2} 
w/ Obj & \multicolumn{1}{c|}{w/ Sup} &   & &  \\
\hline
&  & & 63.2 & 40.1 \\  
\checkmark &  &  & 63.3 & 40.9 \\  
\checkmark & \checkmark &  & 63.2 & 42.0 \\  
 &  & \checkmark & 64.5 & 51.1 \\
\checkmark &  & \checkmark & 65.5 & 52.9 \\ 
\checkmark & \checkmark & \checkmark & 66.2 & 53.5 \\ 

\hline
\end{tabular}}
\caption{Effects of supervising objectness scores (Obj) and suppressing offset predictions (Sup) in the noise suppression module (FSM). PRM denotes the proposal refinement module. The experiment is based on VoteNet re-implemented by MMDetection3D.}
\label{objectness}
\end{table}

\paragraph{Multi-Resolution RoI Pooling.} We use grid-level max-pooling at multiple resolutions to extract features of proposal boxes. As shown in Table \ref{pooling}, extracting features at multiple resolutions is beneficial, especially for accurate locating. Compared with setting resolution $r$ to $5$, using multiple resolutions improves 1.1 mAP@0.5 with negligible budgets.

\begin{table}[tbp]
\centering
\resizebox{0.75\columnwidth}!{\begin{tabular}{ccc|cc|c}
\hline
r=1 & r=3 & r=5 & mAP@0.25   & mAP@0.5 & Time \\
\hline
\checkmark &  & & 64.3 & 47.6 & 67 \\  
 & \checkmark &  & 65.8 & 50.7 & 69 \\  
 & & \checkmark & 66.1 & 52.4 & 70 \\  
\checkmark & \checkmark  & \checkmark & 66.2 & 53.5 & 74 \\  
\hline
\end{tabular}}
\caption{Effect of capturing RoI features at resolution $r$ in RoI pooling. The inference time (ms) is measured on a single NVIDIA Tesla V100.}
\label{pooling}
\end{table}

\paragraph{Feature Gating and Attention.} In the feature gating, we use features from VoteNet to filter RoI features at multiple resolutions separately and sum them up at the end. As shown in Table \ref{gating}, introducing feature gating improves the performance. Besides, using dot multiplication instead of matrix multiplication in the feature gating performs better. Then, we use a transformer layer to model relationships between features. As shown in Table \ref{gating}, removing the transformer layer leads to a significant drop in performance, i.e., 6.7 mAP@0.25 and 5.8 mAP@0.5. However, the performance will improve a lot if we use NMS to suppress redundant predictions. It means that the transformer layer plays an important role in ensuring producing unique predictions. This phenomenon can also be well observed in Figure \ref{refine_merge}. After refining proposals multiple times, the redundant prediction boxes are gradually eliminated. We also notice that after using the transformer layer, further adopting the NMS will cause performance degradation on mAP@0.5.

\begin{table}[tbp]
\centering
\resizebox{0.99\columnwidth}!{\begin{tabular}{cccc|cc}
\hline
w/ GAT & w/ DOT & w/ TF & w/ NMS & mAP@0.25   & mAP@0.5  \\
\hline
 &  & \checkmark & & 65.1 & 51.3  \\  
\checkmark & & \checkmark & & 65.7 & 53.0   \\  
\checkmark & \checkmark & \checkmark & & 66.2 & 53.5  \\
\checkmark & \checkmark & \checkmark & \checkmark  & 66.6 & 52.3  \\
\hline
\checkmark & \checkmark & & & 59.5 & 47.7  \\  
\checkmark & \checkmark & & \checkmark & 64.1 & 49.5  \\  
\hline
\end{tabular}}
\caption{Effect of feature gating (GAT), dot multiplication (DOT), transformer layer (TF), and NMS on SRDet.}
\label{gating}
\end{table}

\paragraph{Proposal Refinement.} We refine prediction boxes multiple times in training and inference. To figure out its effect, we adjust refinement times in our ablation study. As shown in Table \ref{refine}, refining $3$ times achieves the best speed-accuracy trade-off. Besides, not sharing parameters leads to lower performances. We believe that the small size of the dataset is an important factor. We also show the prediction boxes during refinement in Figure \ref{refine_merge} for a better illustration. We can see that most of the objects are well located in the first or second stages, but there is still a certain degree of detection redundancy and false positive. Through multiple stages of proposal interaction, the model can produce more accurate 3D boxes while suppressing redundancy.

\begin{table}[!t]
\centering
\resizebox{0.85\columnwidth}!{\begin{tabular}{cc|cc|c}
\hline
\#Refine & w/ Share & mAP@0.25   & mAP@0.5 & Time \\
\hline
1 & \checkmark & 62.3 & 50.1 & 63 \\ 
2 & \checkmark & 64.9 & 51.4 & 69 \\ 
3 & \checkmark & 66.2 & 53.5 & 74 \\ 
4 & \checkmark & 66.2 & 51.3 & 80 \\ 
\hline
3 & & 64.8 & 51.8 & 75 \\
\hline
\end{tabular}}
\caption{Precision and inference time (ms) when adjusting the refinement number and removing the parameter sharing.}
\label{refine}
\end{table}

\begin{figure*}[!t]
\centering
\includegraphics[width=1.87\columnwidth]{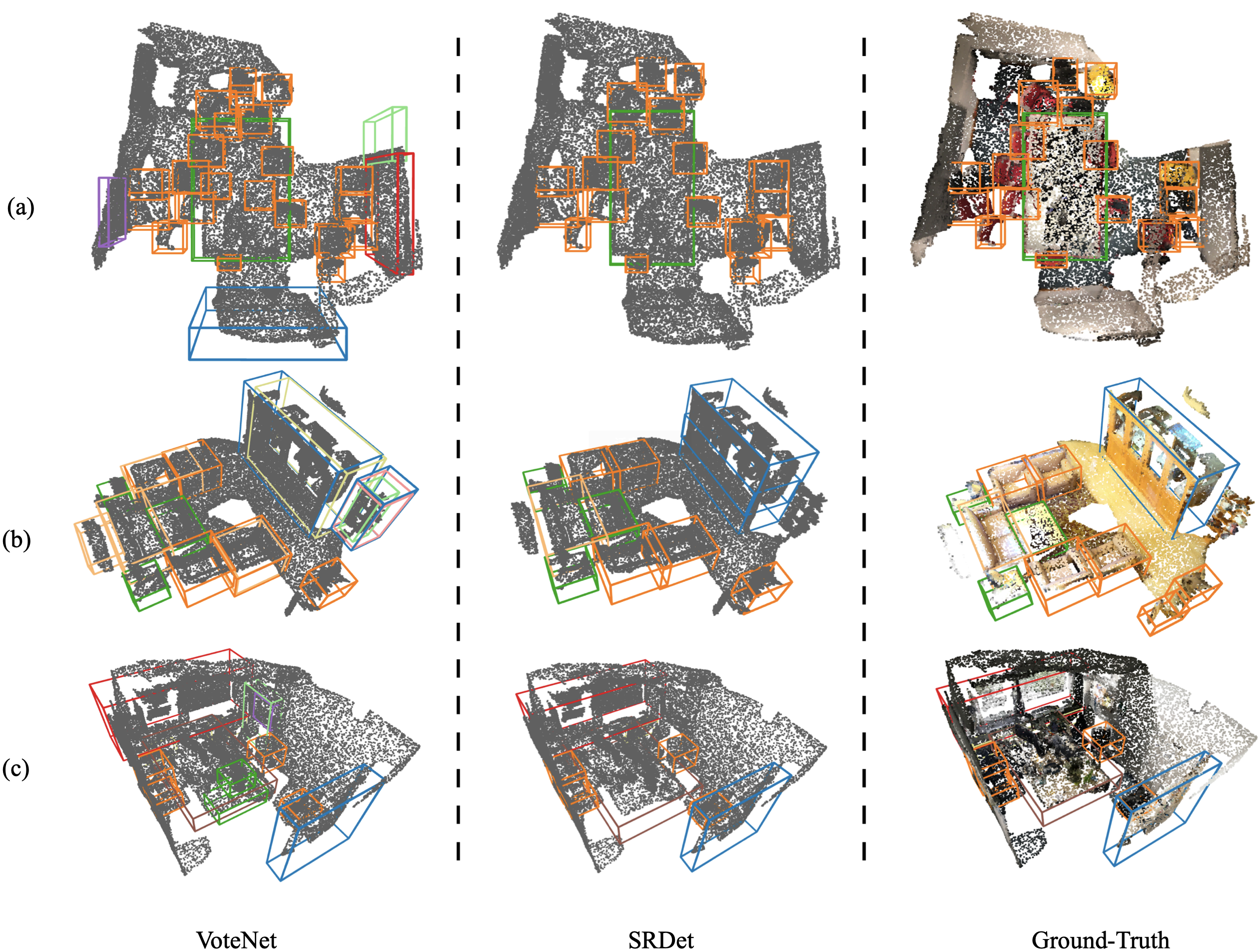}
\caption{Qualitative results of 3D object detection on the ScanNetV2.}
\label{q_results}
\end{figure*}


\section{Qualitative Results and Discussion}
We compare the qualitative results of VoteNet and SRDet on the ScanNetV2 validation set in Figure \ref{harm} and Figure \ref{q_results}. We can observe that SRDet can suppress those scattered vote points and significantly reduce false positives in results, respectively. In particular, we can see in Figure \ref{q_results} (b) and (c) that VoteNet tends to produce detection results with different categories for the same instance, which are difficult to filter out by using commonly adopted class-aware NMS (NMS is only performed within objects of the same class.). Besides, the class-agnostic NMS also cannot handle this situation well. For instance, Figure \ref{q_results} (a) shows that the table and the chair overlap, and the class-agnostic NMS may filter out the chair or the table and cause false negatives. In contrast, benefiting from the proposal refinement module, SRDet can generate a certain type of detection result for a single instance without resorting to NMS, and can handle occlusion and non-occlusion conditions well. This advantage makes SRDet a more practical detector in the real-world scenario.

\section{Conclusion}
We introduce SRDet, the first fully end-to-end 3D object detector building on VoteNet. We focus on the two main challenges: geometric point noise caused by the disordered and varying-density nature of point clouds and the substantial computation overhead required by the 3D space. Therefore, we present the suppress-and-refine framework to efficiently noise points and feed them to proposal boxes for the following RoI-aware refinement. We show that noise points can be well suppressed with the proposed noise suppression module, and the proposal refinement module can produce high-quality results with a small computation budget. Our SRDet achieves state-of-the-art performances on ScanNetV2 and SUN RGB-D while keeping a fast inference speed, which demonstrates that the proposed method is effective and works well in the 3D domain.

{\small
\bibliographystyle{ieee_fullname}
\bibliography{egbib}
}

\newpage
\begin{appendices}

\renewcommand\thesection{\Alph{section}}
\renewcommand\thesubsection{\Alph{section}{.}\arabic{subsection}}
\section{More Ablation Studies}

\subsection{Loss Weight.} In our implementation, the loss weight of the proposal refining module is set as $w_{cls}=1.5$, $w_{L1}=0.45$, $w_{iou}=2$, $w_{cor}=0.25$. As shown in Table \ref{lossw}, the detection accuracy is not sensitive to the choice of the loss weight.

\subsection{Proposal Number.} We set the proposal number $k=160$ in VoteNet and sample $K=128$ boxes according to their prediction scores. As shown in Table \ref{proposalnum}, both proposal sampling and enlarging the proposal number improve the performance.

\begin{table}
\centering
\resizebox{0.75\columnwidth}!{\begin{tabular}{cccc|cc}
\hline
$w_{cls}$ & $w_{L1}$ & $w_{iou}$ & $w_{cor}$ & mAP@0.25 & mAP@0.5 \\
\hline
 1.0 & 0.45 & 2.0 & 0.25 & 64.1 & 52.4 \\  
 2.0 & 0.45 & 2.0 & 0.25 & 67.2 & 52.0 \\  
 \hline
 1.5 & 0.3 & 2.0 & 0.25 & 66.3 & 51.7 \\  
 1.5 & 0.6 & 2.0 & 0.25 & 65.6 & 53.0 \\  
 \hline
 1.5 & 0.45 & 1.0 & 0.25 & 66.5 & 52.1 \\  
 1.5 & 0.45 & 3.0 & 0.25 & 64.9 & 53.4 \\  
 \hline
 1.5 & 0.45 & 2.0 & 0.1 & 65.7 & 52.4 \\  
 1.5 & 0.45 & 2.0 & 0.4 & 66.3 & 52.7 \\  
 \hline
 1.5 & 0.45 & 2.0 & 0.25 & 66.2 & 53.5 \\  
\hline
\end{tabular}}
\caption{Ablation study on the loss weight. }
\label{lossw}
\end{table}

\begin{table}
\centering
\resizebox{0.65\columnwidth}!{\begin{tabular}{cc|cc|c}
\hline
$k$ & $K$ & mAP@0.25 & mAP@0.5 & Time \\
\hline
 256 & 128 & 65.4 & 53.1 & 74 \\  
 256 & 160 & 66.7 & 53.0 & 75 \\  
 256 & 192 & 67.8 & 54.0 & 76 \\  
 256 & 256 & 66.0 & 53.1 & 88 \\  
 160 & 160 & 65.0 & 52.6 & 75 \\  
 \hline
 160 & 128 & 66.2 & 53.5 & 74 \\  
\hline
\end{tabular}}
\caption{Ablation study on the proposal number. The inference time (ms) is measure on single NVIDIA Tesla V100.}
\label{proposalnum}
\end{table}
\end{appendices}

\end{document}